\documentclass[conference]{IEEEtran}
\usepackage{cite}
\ifCLASSINFOpdf
 \usepackage[pdftex]{graphicx}
\else
\fi
\usepackage{epstopdf}

\usepackage{amsmath}
\usepackage{algorithmic}
\usepackage{hhline}

\usepackage{multirow}
\usepackage{enumitem}

\usepackage{url}
\begin{document}
\onecolumn
\copyright 2019 IEEE. Personal use of this material is permitted. Permission from IEEE must be obtained for all other uses, in any current or future media, including reprinting/republishing this material for advertising or promotional purposes, creating new collective works, for resale or redistribution to servers or lists, or reuse of any copyrighted component of this work in other works. 
\\
DOI: 10.1109/ISGTEurope.2019.8905453

\newpage
%\twocolumn
\title{A Machine Learning Model for Long-Term Power Generation Forecasting at Bidding Zone Level}

\author{\IEEEauthorblockN{Michela Moschella, Mauro Tucci, Emanuele Crisostomi}
\IEEEauthorblockA{Department of Energy, Systems, Territory \\
and Constructions Engineering, University of Pisa\\
Pisa, Italy \\
michela.moschella@ing.unipi.it}
\and
\IEEEauthorblockN{Alessandro Betti}
\IEEEauthorblockA{i-EM s.r.l.\\
Livorno, Italy\\
alessandro.betti@i-em.eu}}

%\IEEEoverridecommandlockouts
%\IEEEpubid{\begin{minipage}[t]{\textwidth}\ \\[10pt]
%        \centering\normalsize{978-1-5386-8218-0/19/\$31.00 \copyright 2019 IEEE}
%\end{minipage}} % see https://tex.stackexchange.com/questions/454280/how-to-center-ieee-copyright-at-the-bottom-of-the-first-page

\maketitle

\begin{abstract}
The increasing penetration level of energy generation from renewable sources is demanding for more accurate and reliable forecasting tools to support classic power grid operations (e.g., unit commitment, electricity market clearing or maintenance planning). For this purpose, many physical models have been employed, and more recently many statistical or machine learning algorithms, and data-driven methods in general, are becoming subject of intense research. While generally the power research community focuses on power forecasting at the level of single plants, in a short future horizon of time, in this time we are interested in aggregated macro-area power generation (i.e., in a territory of size greater than 100000 km$^2$) with a future horizon of interest up to 15 days ahead. Real data are used to validate the proposed forecasting methodology on a test set of several months.
\end{abstract}

% \fi
\IEEEpeerreviewmaketitle

\section{Introduction}

\subsection{Motivations}
As the penetration level of Renewable Energy (RE) sources is growing worldwide to meet ever tightening sustainability goals \cite{ren21}, the intermittent and uncertain nature of RE is posing increasing challenges to efficiently manage a power grid, eventually endangering its own stability. In this context, the availability of accurate forecasts of power generation from RE may mitigate the impact of the increasing penetration level and improve the operation of power systems \cite{bibref:gigoni}. 
In particular, in this paper we are interested in developing long-term RE power generation forecasting algorithms, up to 15 days ahead for aggregated areas. Such a long horizon of time ahead is convenient for maintenance scheduling, for planning tactic upgrades or for planning switching on/off of big conventional plants when future power generation from RE is expected to be particularly low or high, with respect to the load. In addition, we focus on forecasting algorithms operating at aggregated level, where a region here is a bidding zone in Italy (see Fig.~\ref{fig:map}). We aggregate data in terms of bidding zones since many measured and predicted data are available for this aggregation level (e.g., load consumption predicted and measured data are available at the Terna (Italian TSO) website\footnote{https://www.terna.it/en-gb/sistemaelettrico/transparencyreport/load.aspx}).

\subsection{State of the Art}
There is a rich literature on power generation from RE, usually differing in terms of the future horizon of prediction that may range from $1$ second to $6$ hours,  from $6$ hours up to day ahead, and from $2$ days ahead or longer, which correspond to the \emph{intra-day}, the \emph{day-ahead} and the \emph{long-term} models, respectively. In this case a general overview of existing forecast methods can be found in \cite{bibref:rev_pv} and \cite{bibref:rev_wd} for Photovoltaic (PV) and Wind (WD) power generation, respectively.\\
We can also distinguish models with different data geographical resolution; the most popular spatial aggregation is the \emph{pointwise} one, where data refer to a single power plant, but there is also a type of forecasting regarding regions and macro-areas; in the last case, the problem is completely different since the size and the location of all plants in a given region are not generally known. \\
Very few works make forecasts at large-scale regional areas; most focus on power generation from single PV or wind plants, that are successively aggregated following the so-called  \emph{up-scaling} method \cite{IEA2013}. A similar paper is \cite{bibref:pierro}, where however only the PV case is considered, and for much smaller scales than in our case and only up to $2$ days ahead. Another classic paper on this topic is \cite{bibref:marinelli}, that however did not use information on energy production from Renewable Energy Sources (RES). This piece of information is on the other hand available in our case study with hourly resolution, thus allowing us to propose more accurate forecasting models. 

\subsection{Contributions}
While plenty of papers have been written for power generation from RE, our problem is rather a peculiar one because we are trying to predict power generation in wide areas (in some cases greater than $100000$ km$^2$) for a long horizon of time. The problem is particularly challenging since we do not know where power generation plants are exactly located in such areas, nor their nominal size. Thus, pure data-driven methodologies, namely k-Nearest Neighbous (k-NN) and Quantile Regression Forest (QRF), are used to predict power generation on the basis of large historical data-sets. \\
While not too accurate results are obtained, especially for a horizon of forecast larger than $5$ days, still the forecasting results may be accurate enough to support classic power grid operations (e.g., maintenance planning).

\section{Data set description}
In our problem we want to forecast power generated from solar and wind sources at an aggregated level (i.e. Italian bidding zones), using meteorological data as input variables (see Table~\ref{tab:vars} for a brief summary of used variables).

\subsection{Power generation data}\label{subsec:power}
\begin{figure}[!t]
\centering
\includegraphics[width=2in]{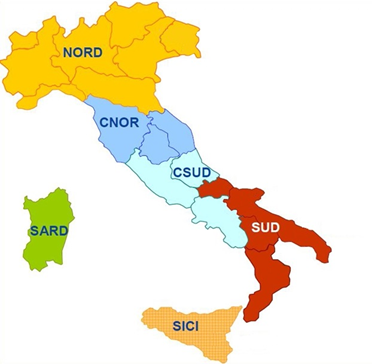}
\caption{Bidding zones of Italy, identified by Terna.}
\label{fig:map}
\end{figure}

Power generation data are available from the aforementioned Terna website\footnote{\url{https://www.terna.it/SistemaElettrico/TransparencyReport/Generation/Expostdataontheactualgeneration.aspx}}. 
In particular, we have at our disposal \emph{hourly data} for each Italian bidding zone (indicated with acronyms NORD, CNOR, CSUD, SUD, SICI and SARD, as shown in Fig.~\ref{fig:map}), where the hourly value is the average power over the previous hour.

The previous data are known to contain some inaccurate information. In the pre-processing stage illustrated in Sec.~\ref{sec:methods} we also use the monthly power generation data (aggregated at national level) that is known to be more accurate.

%[ If I find some official report about the nature of PV and wind data, I will add some information about it. Actually, we know informally that the PV data are the results of a summation between measured and estimated values]

\subsection{Meteorological data}\label{sec:metdata}
For the prediction of power generated from PV plants, we exploit the hourly satellite Global Horizontal Irradiance (GHI) and Global Horizontal Irradiance in Clear Sky conditions (GHI$_{CS}$) from $2015$ on, provided by Flyby s.r.l.\footnote{http://www.flyby.it/} (for more details on data validation see~\cite{bibref:flyby}), as well as the forecasting meteorological GHI provided by Aeronautica Militare (AM)\footnote{http://www.meteoam.it/modelli-di-previsione-numerica} from 2016 onwards. For WD farms we use the two components of the wind speed, i.e. the west-to-east component (UGRD), and the south-to-north one (VGRD). 
%Due to the lack of measured data, we used the 24-hour ahead hourly forecast of UGRD and VGRD provided by AM. 

\textbf{Remark: } Forecast meteorological data provided by AM come from two different models, depending on the forecast horizon they refer to. In particular,
\begin{itemize}
\item data referred to the forecast interval $[+0h, +72h]$, are outputs of the model \emph{COSMO-ME}\footnotemark[\value{footnote}], a local model on the south-central Europe and Mediterranean Sea;
\item for data about the horizon $[+75h, +240h]$, the global IFS model\footnote{https://www.ecmwf.int/en/forecasts/datasets/set-i} (of ECMWF) is adopted.
\end{itemize}
Unfortunately, available forecast data do not cover the whole forecast horizon of $15$ days, and consequently we have to fill such missing values; a simple \emph{persistence} technique is used (i.e., the forecast of the last day is kept constant for the following days as well).

As the previous meteorological data (available in a \emph{raster} format) cover wide areas, to compress the size of the input data, we aggregate the meteorological data to the level of Provinces ($110$ in Italy).

\begin{table}[htbp]
\caption{MODEL VARIABLES}
\begin{center}
	\begin{tabular}{|c|c|c|c|}
	\hline
	\textbf{Module}&\textbf{Variable}&\textbf{Time resolution}& \textbf{Spatial resolution} \\
	\hline
	\multirow{2}{*}{PV}&GHI& 1 hour & Province \\
       \cline{2-4}
	 &GHI$_{CS}$& 1 hour & Province \\
	 \cline{2-4}
       &PV generated power& 1 hour & bidding zone \\
	 \cline{2-4}
       & PV generated power&1 month&Italy \\
	\hhline{|=|=|=|=|}
	\multirow{2}{*}{WD}&UGRD& 1 hour & Province\\
	\cline{2-4}
	 &VGRD& 1 hour & Province \\
	\cline{2-4}	 
	&WD generated power& 1 hour & bidding zone \\
	\cline{2-4}
	& WD generated power&1 month&Italy \\
	\hline
%	\multicolumn{4}{l}{\multirow{2}{*}{$^{\mathrm{a}}$Power refers to both hourly and monthly power data, with aggregation}}\\
%	
%	    \multicolumn{4}{l}{at bidding zone and Italy level, respectively.} 
	\end{tabular}
\label{tab:vars}
\end{center}
\end{table}

\section{Methods} \label{sec:methods}
Following the notation of \cite{Breiman2001}, we group variables into \emph{predictor} and \emph{response} variables classes. 
As in a \emph{black box} identification procedure, we use machine learning algorithms to identify an unknown and most likely, non-linear, function that maps the predictors vector $x$ into the response variable $y$, i.e. 
\begin{equation}\label{enq:yfx}
y = f(x).
\end{equation}
In our case, predictors correspond to the meteorological variables described in Sec. \ref{sec:metdata}, while the generated power is the response variable of the model. 
We use two models, one for PV and one for WD power generation forecasting, and in both cases data are divided in two parts: a \emph{training set} with measured data, used to learn Eq.~\eqref{enq:yfx}, and the \emph{test set} with forecast data, exploited for evaluation and comparison.

\subsection{Preprocessing and Missing data}
A preprocessing phase is fundamental to remove outliers and prevent an identification methodology from learning wrong patterns in the training set \cite{Engel2007}. \\
First, the hourly values of the power time-series are proportionally scaled using the more accurate total national monthly generation. Additionally, for PV instances, outliers are identified as those irradiance-power pairs that fall out of a safety cone in the $(\overline{GHI}, P)$ plane, where $\overline{GHI}$ is the average satellite irradiance over each bidding zone, and $P$ is the related hourly power, inspired by the procedure outlined in \cite{fuentes}.

Very few data are missing in our data-set. Cubic spline interpolation is used to fill such gaps.

\subsection{Machine Learning algorithms}
A cooperative \emph{ensemble} of \emph{k-NN} and \emph{QRF} methods is used to forecast power generation. \\

\subsubsection{k-NN}

k-NN is a parametric method based on the assumption that a given weather forecast is most likely to yield a power generation close to past instances with similar weather conditions. The $k$ most similar samples in the past are selected and the corresponding historical powers are combined, with a weight depending on the similarity degree. Parameters of the k-NN are the number $k$ of nearest neighbours, the distance metric, and the kernel used for weights modeling. In our case, Euclidean metric is used for the distance and hyperbolic kernel is exploited for the combination of $k$ nearest neighbours \cite{bibref:gigoni}. \\
Additionally, for the PV case we exploit the periodicity in the irradiance time series by selecting in the past the $k$ nearest neighbours of a similar hour and month of the specific power sample to be predicted.
In particular, if $M$ and $H$ are the month and the hour of the test sample to be predicted, respectively, only training instances with month and hour \emph{close} to $M$ and $H$ are selected. \\
A summary of inputs used for the algorithm is shown in Table ~\ref{tab:input}.

\subsubsection{QRF}
The QRF algorithm is a variant of \emph{Random Forest} (RF) developed by Breiman \cite{bibref:breim} which, unlike conventional RF, takes track of all the target samples and not just their average. RF is a collection of decision trees that are combined together to enhance the predictive capability of a single tree, by approximating the mean conditional distribution of the response variable. On the other hand, QRF \cite{bibref:qrf} provides the full conditional distribution of the response variable.
In particular, assuming that $Y$ is the target variable and $X$ the vector of predictors, then the final goal is finding the relationship between $X$ and $Y$. A conventional RF estimates the conditional mean of the target $Y$, given the attribute $X=x$. Instead the QRF, given a certain level $\alpha$ ($0 < \alpha < 1$), estimates the conditional quantile
\begin{equation}\label{eqn:quantile}
Q_\alpha (x) = \inf \{ y : F(y| X=x) \ge \alpha\},
\end{equation}
where $F(y|X=x)$ is the conditional distribution function of $Y$ given $X=x$. Consequently, QRF is a substantial improvement of conventional RF, because the $\alpha$-quantile $Q_\alpha (x)$ gives a more complete information about the distribution of $Y$ than the output of the conditional mean provided by RF \cite{bibref:qrf}. \\
The input variables used for this algorithm are shown in Table ~\ref{tab:input}.

\begin{table}[htbp]
\caption{Inputs of the k-NN and QRF algorithms}
\begin{center}
\begin{tabular}{|c|c|c|}
\hline
\textbf{Algorithm}&\textbf{PV}&\textbf{WD} \\
\hline
\textbf{k-NN}& GHI, GHI$_{CS}$& wind speed \\
\hline
\textbf{QRF}&GHI, GHI$_{CS}$, Month, Hour& UGRD, VGRD\\
\hline

\end{tabular}
\label{tab:input}
\end{center}
\end{table}

\subsection{PV model: post-processing phase}\label{sec:postp}
Only in the PV model, we post-process the k-NN and QRF before combining them in an ensemble. \\
The rationale for doing so is the dependence of power generation on year and season, so same irradiance values may correspond to different generated powers, at different time of the years.\\
The main steps of post-processing are the following:
\begin{enumerate}
\item get the mean ratio $Q_{for}$ between the power generation forecasts referring to peak values of GHI and the related GHI values, in the interval of prediction (e.g. $360$ h);
\item compute the mean ratio between measured power and irradiance data (corresponding to peak values of GHI) of the $n$ weeks before the test set in training set ($Q_{train}$);
\item finally compute the factor $K_{prod} = \frac{Q_{train}}{Q_{for}}$ and scale the prediction of the generated power by the same factor.
\end{enumerate}
It is worth remarking that this rescaling is adequate only if the training set is immediately preceding the test period, otherwise the ratio $Q$ would change and such post-processing would not be convenient.

\subsection{Model tuning}
Some hyperparameters need to be tuned for the procedure, such as:
\begin{itemize}
\item the threshold for removing outliers in the bi-variate pre-processing phase of the PV module;
\item the level $q$ of the quantile in the QRF algorithm;
\item the number $k$ of nearest neighbours in the k-NN algorithm;
\item the number $n$ of weeks in the post-processing of Sec.~\ref{sec:postp}.
\end{itemize}
In our case study we use a trial and error method on the training set.

\section{Results and Discussion}
We now validate our model performances on a test period of $6$ months, from January to June 2017. We adopt a \emph{semi-moving} window technique to select the training period: the start date is fixed ($1$st May $2015$), whereas the end date is the day before the test period.\\
We evaluate the performances every month by considering $2$ different error metrics: the Normalized Mean Bias Error (NMBE) and the Normalized Root Mean Squared Error (NRMSE).
If we denote by $\hat{y}$ the prediction of our model, and by $y$ the actual generated power, if $e_i=\hat{y}_i - y_i$ is the forecasting error of the $i$-th hour within the horizon of forecast, then the error metrics are defined as:
\begin{enumerate}[label=(\roman*)] 
\item \label{eq1}$NMBE= \frac{1}{N}\sum_{i=1} ^N \frac{e_i}{M_m}; $
\item \label{eq2} $NRMSE =  \sqrt{\frac{1}{N} \sum_{i=1} ^N \Big(\frac{e_i}{M_{m}}\Big) ^2}$, 
\end{enumerate}
where $M_{m}$ was used for normalization purposes and was chosen as the historical maximum power value $M$ in the month $m$.  
In addition, the number $N$ is equal to the number of items considered for the computation of the scores; for instance, if we want to compute the error referred only to forecasts of 1 day ahead, we will select all the items in the test month $m$ related to this forecast horizon (i.e., $N$ will be equal to the number of days in the month multiplied by $24$ (number of hourly values in 1 day)).
%Every month, $N$ was equal to the number of days of the month times the number of hours of the forecasting horizon (here, $360$ hours corresponding to $15$ days). 

Since these two metrics provide normalized values, they allow a comparison between different bidding zones (with different installed nominal power). In addition, they also highlight specific characteristics of the model performances; 
actually, the NMBE metric provides information about the error polarization (i.e., if the model was overestimating or underestimating observed values), whereas the NRMSE takes account of the absolute error, avoiding balancing effects due to the pointwise errors signs.

%In Fig.~\ref{fig:error_horizon} the NRMSE is displayed as a function of the forecast horizon, from one day ahead up to 15 days ahead, averaged over bidding zones. As expected, the error increases by almost $10$\%($19$\%) from 3 days ahead up to $79$\%($160$\%) 15 days ahead with respect to day $+1$ for PV(WD) model, where the mean error is about $7.27$\%($11.23$\%).
\subsection{NMBE and NRMSE: errors analysis}
\begin{figure}[!t]
\centering
\includegraphics[width=2.9in]{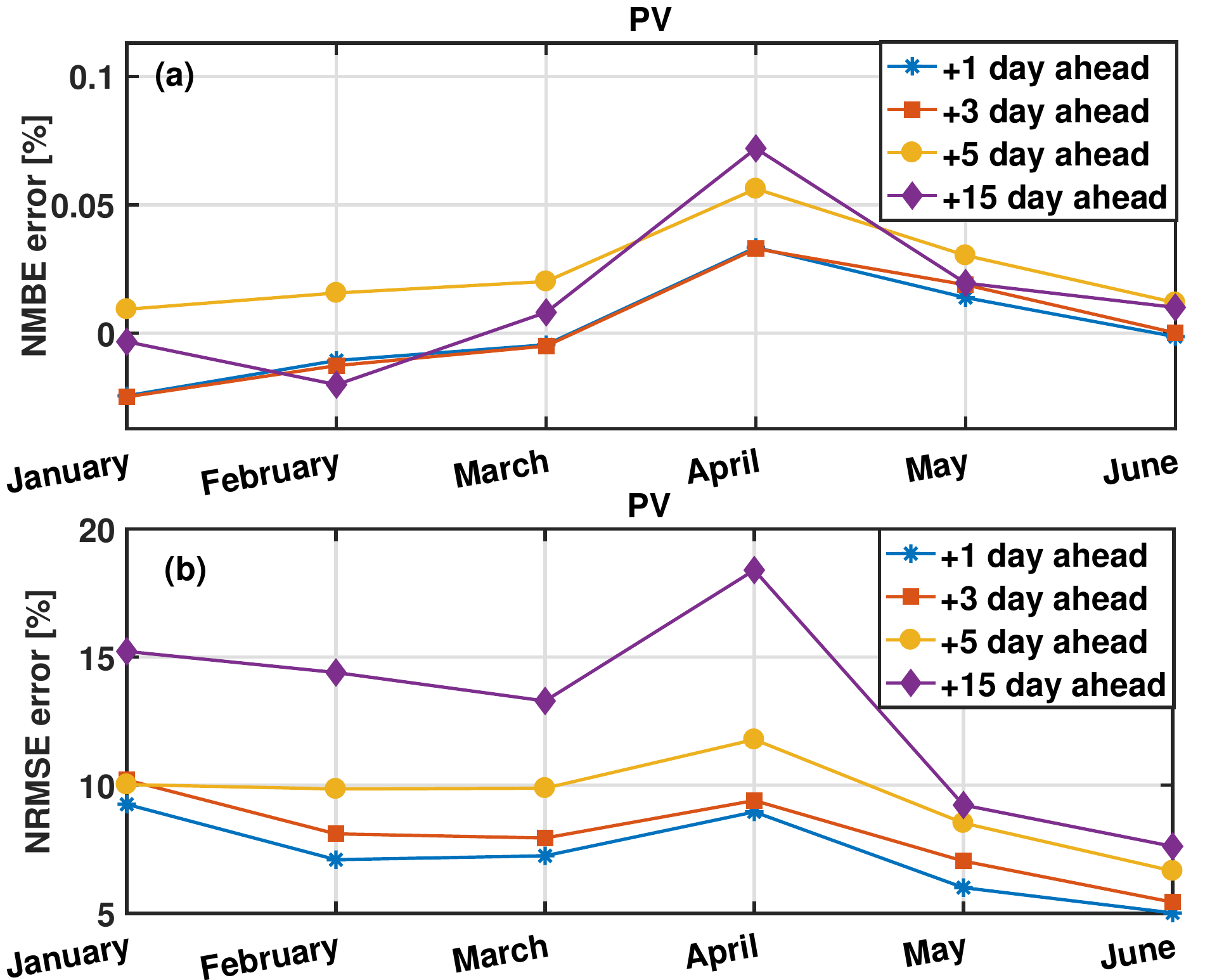}
\caption{NMBE (a) and NRMSE (b) averaged over bidding zones as a function of test month for PV plants.}
\label{fig:it_pv}
\end{figure}

\begin{figure}[!t]
\centering
\includegraphics[width=2.9in]{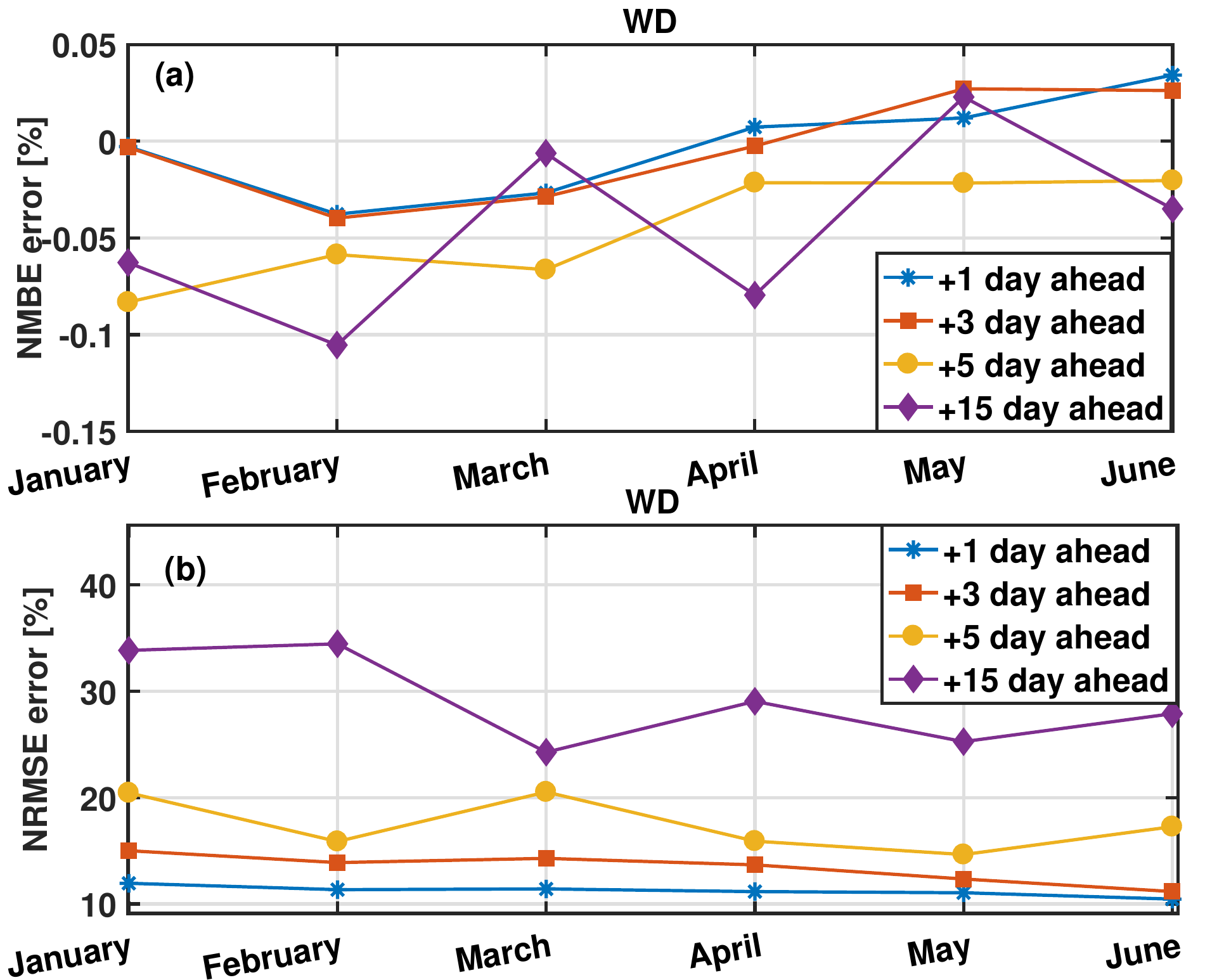}
\caption{NMBE (a) and NRMSE (b) averaged over bidding zones as a function of test month for WD farms.}
\label{fig:it_wd}
\end{figure}

Fig.~\ref{fig:it_pv} and Fig.~\ref{fig:it_wd} show the values of \ref{eq1}-\ref{eq2} errors as a function of the test month at national level, from one-day ahead up to 15 days ahead (for PV and WD power generation, respectively).\\
\newline
The two metrics NMBE and NRMSE emphasize two different aspects of the forecasting error. In particular, the NMBE allows one to appreciate the sign of the error, and to infer whether there is a constant bias in the provided forecasts (i.e., whether the provided forecasts systematically underestimate, or overestimate, power generation). At this regard, both Fig.~\ref{fig:it_pv} and Fig.~\ref{fig:it_wd} show that the error may take any of the two signs in different cases, with rare significant underestimates (the most evident occurs in the wind case in the month of February for the 15 days ahead forecast) or overestimates (the most evident occurs in the PV case in the month of April, for the 15 days ahead forecast again).\\
\newline
On the other hand, the NRMSE plots allow one to better understand the magnitude of the forecasting error. Several results can be appreciated in this case: first of all, as one would expect, the error systematically increases with the length of the future horizon of forecasting. In particular, the error ranges from 5\% (1-day ahead) to 18\% (15 day-ahead) in the PV case, and from 10\% (1-day ahead) up to 35\% (15-day ahead) in the WD case. In addition, one may note that the error is pretty much constant in the wind case (Fig.~\ref{fig:it_wd}(b)) while seasonality effects can be easily identified in the PV case (Fig.~\ref{fig:it_pv}(b)), as weather forecasts are uncertain in spring time, while it is simpler to make irradiance forecasts in Italy in June (most likely, it is going to be a sunny day). Finally, it is much simpler to predict irradiance than wind speed, as the NRMSE is much lower in the PV case than it is in the WD case.

\subsection{Bidding zones and Italian country}
\begin{figure}[!t]
\centering
\includegraphics[width=3.2in]{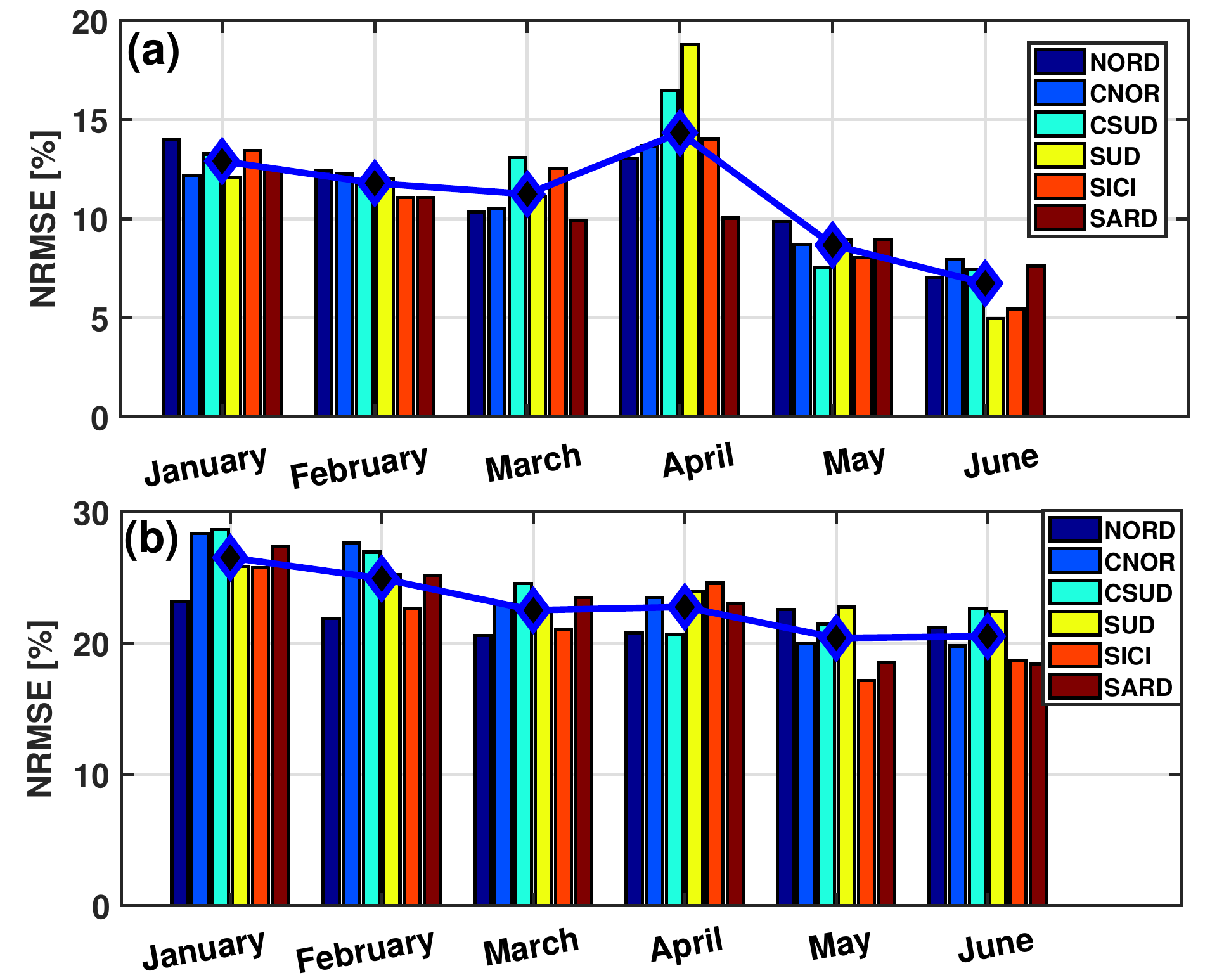}
\caption{NRMSE as a function of test month for PV (plot (a)) and WD (plot (b)) power generation of different bidding zones, computed on the whole forecast horizon of 15 days. The mean NRMSE at Italian level is also shown as a function of time (blue diamonds).}
\label{fig:bz}
\end{figure}

Here NRMSE error is studied as a function of the test month separately for each bidding zone. \\
Fig.~\ref{fig:bz} shows the error computed on the whole forecast horizon of 15 days in order to gain a global performance indicator. 
It is worth noticing that usually the trend of each single macro-area followed the overall trend at national level, identified by the blue curve, but some deviations may occur, as it can be seen for example in Fig.~\ref{fig:bz}(a) for the month of April, where SUD exhibits a score $30$\% higher than the average global error (about $14.34$ MW). 
Negative deviations may also occur, as it can be observed for example in Fig.~\ref{fig:bz}(b) in April for CSUD ($9$\% less with respect to the mean value). 
Such behaviour highlights the dependence of the model performance on the considered bidding area and hints that a benefit may be obtained by tuning the model specifically for different areas (we leave this for future work).

\subsection{Forecasts versus Metering time series}
\begin{figure}[!t]
\centering
\includegraphics[width=3in]{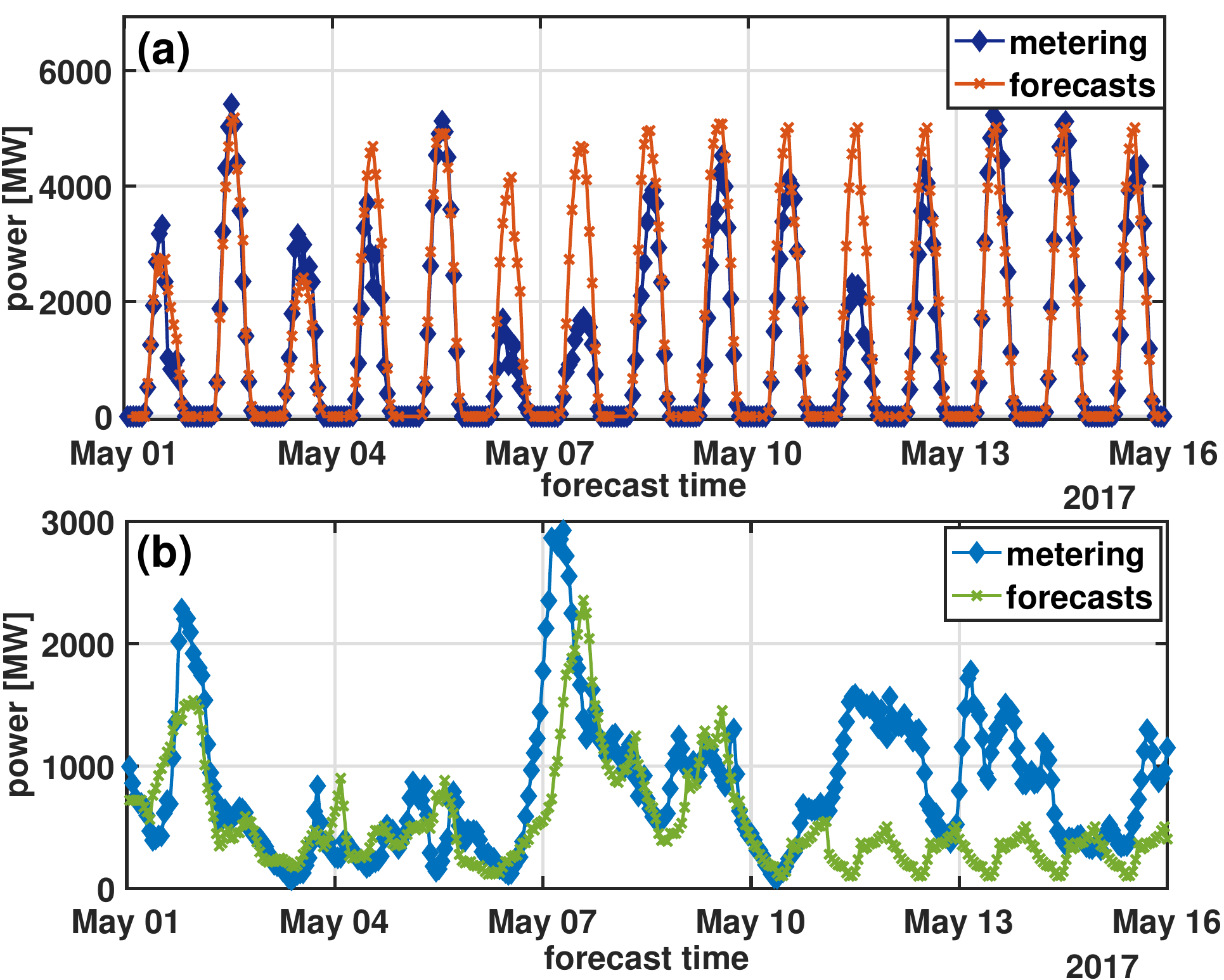}
\caption{(a) PV case: forecasts (orange curve) and metering (blue curve) as a function of time for PV power generation of NORD bidding zone; (b) WD case: forecasts (orange curve) and metering (blue curve) as a function of time for WD power generation of SUD bidding zone. The runtime of the $1st$ of May $2017$ is shown.}
\label{fig:comp}
\end{figure}

In this section we show an example of comparison between our model forecasts and the actual measured power; in particular, Fig.~\ref{fig:comp} shows the results referring to the run of May 1, for both PV and WD power generation, considering the most productive bidding zone (NORD for PV, and SUD for WD \cite{bibref:gse}, respectively). 
As we can see from Fig.~\ref{fig:comp}, a good agreement between the forecast and the actual value can be observed for the first $5$ days ahead, whereas a degradation is observed from the $6$-th day ahead on (after May $6$).
This effect is strongly related to the model dependence on the input accuracy: the larger the forecast horizons, the less reliable the meteorological data are. 
Finally, it should be observed that in the WD module, the last $5$ days were affected by the persistence error, whereas in the PV case such effect was mitigated by the periodicity of the power curve.

\section{Conclusion}
In this paper we provide a power generation forecasting model at large-scale regional areas, where the macro-areas of interest are the Italian bidding zones.\\
Good results are achieved, thanks also to the use of a large historical data-set. More precisely, good performance can be observed for the first $5$ days ahead, and results may be accurate enough to support classic power grid operation.\\
Further study can be however conducted; for instance, the implementation of specific tuning stages for each bidding zone, separately. Finally, another improvement may refer to the PV model; actually, the introduction of Global Tilted Irradiance (GTI) is expected to lead to a performance improvement, and for this reason will be eventually introduced in the model in a future work.

\section*{Acknowledgment}
EC and MM were supported in part by the University of Pisa under Project PRA 2018 38 “Smart Energy Systems”.

%\bibliographystyle{IEEEtran}

\begin{thebibliography}{1}


\bibitem{ren21}Renewable Energy Policy Network for the 21st~Century (REN21),  ``Renewables 2018 Global Status Report," Paris REN21 Secretariat, 2018.

\bibitem{bibref:gigoni}L.~Gigoni, A.~Betti, E.~Crisostomi, A.~Franco, M.~Tucci, F.~Bizzarri, and D.~Mucci, ``Day-Ahead Hourly Forecasting of Power Generation From Photovoltaic Plants," IEEE Transactions on Sustainable Energy, vol. 9, no. 2, pp. 831-842, April 2018.

\bibitem{bibref:rev_pv}J.~Antonanzas, N.~Osorio, R.~Escobar, R.~Urraca, F.~M.~de Pison, and F.~Antonanzas-Torres, ``Review of photovoltaic power forecasting,” Solar Energy, vol. 136, pp. 78-111, 2016. 

\bibitem{bibref:rev_wd}S.~S.~Soman, H.~Zareipour, O.~Malik and P.~Mandal, ``A review of wind power and wind speed forecasting methods with different time horizons," North American Power Symposium 2010, Arlington, TX, 2010, pp. 1-8.

\bibitem{IEA2013}S.~Pelland, J.~Remund, J.~Kleissl, T.~Oozeki, K.~De Brabandere,  ``Photovoltaic and Solar Forecasting: State of the Art," Report IEA PVPS T14‐01:2013, October 2013.

\bibitem{bibref:pierro}M.~Pierro, M.~De Felice, E.~Maggioni, D.~Moser, A.~Perotto, F.~Spada, C.~Cornaro, ``Data-driven upscaling methods for regional photovoltaic power estimation and forecast using satellite and numerical weather prediction data," Solar Energy, vol. 158, pp. 1026-1038, 2017.

%\bibitem{bibref:pierro}M.~Pierro, M.~De Felice, E.~Maggioni, D.~Moser, A.~Perotto, F.~Spada, C.~Cornaro, ``A New Approach For Regional Photovoltaic Power Estimation And Forecast, " 33rd European Photovoltaic Solar Energy Conference and Exhibition, September 2017.

\bibitem{bibref:marinelli}M.~Marinelli, P.~Maule, A.~N.~Hahmann, O.~Gehrke, P.~B.~Nørgård, N.~A.~Cutululis, ``Wind and photovoltaic large-scale regional models for hourly production evaluation", IEEE Transactions on Sustainable Energy, vol. 6, no. 3, pp. 916-923, July 2015.

\bibitem{bibref:flyby}M.~Morelli, A.~Masini, F.~Ruffini, M.~A.~C.~Potenza, ``Web tools concerning performance analysis and planning support for solar energy plants starting from remotely sensed optical images," Environmental Impact Assessment Review, vol. 52, pp. 18-23, 2015.

\bibitem{Breiman2001}L.~Breiman, ``Statistical Modeling: The Two Cultures,'' Statistical Science, vol. 16, no. 3, pp. 199-215, 2001.

\bibitem{Engel2007}A.~P.~Engelbrecht, Computational Intelligence: An Introduction, 2nd ed., John Wiley~\&~Sons Inc, 2007.

\bibitem{fuentes}M.~Fuentes, G.~Nofuentes, J.~Aguilera, D.L.~Talavera, M.~Castro, ``Application and validation of algebraic methods to predict the behaviour of crystalline silicon PV modules in Mediterranean climates," Solar Energy, vol. 81, no. 11, pp. 1396-1408, Nov. 2007.
%\bibitem{Skopla}E.~Skoplaki, J.A.~Palyvos, ``On the temperature dependence of photovoltaic module electrical performance: A review of efficiency/power correlations," Sol. Energy 83 (5) (2009) 614-624.

\bibitem{bibref:breim} L.~Breiman, ``Random forests,” Machine Learning, vol. 45, no. 1, pp. 5-32, 2001. 

\bibitem{bibref:qrf}N.~Meinshausen, ``Quantile regression forests," The Journal of Machine Learning Research, vol. 7, pp. 983-999, 2006. 

%\bibitem{bibref:irradiance}M.~Diagne, M.~David, P.~Lauret, J.~Boland, N.~Schmutz,``Review of solar irradiance forecasting methods and a proposition for small-scale insular grids," Renewable and Sustainable Energy Reviews, vol. 27, pp. 65-76, 2013.

\bibitem{bibref:gse}GSE~S.p.A., ``Rapporto Statistico, Energia da fonti rinnovabili in Italia, Anno 2016," Gennaio 2018.

%\bibitem{b3} I. S. Jacobs and C. P. Bean, ``Fine particles, thin films and exchange anisotropy,'' in Magnetism, vol. III, G. T. Rado and H. Suhl, Eds. New York: Academic, 1963, pp. 271--350.

\end{thebibliography}

\end{document}